\documentclass[runningheads]{llncs}
\usepackage{graphicx}

\usepackage{cite}
\usepackage{amsmath,amssymb,amsfonts}
\usepackage{algorithmic}
\usepackage{textcomp}
\usepackage{textgreek}
\usepackage{mathtools}
\usepackage{wrapfig}
\usepackage{hyperref}
\usepackage{fancyhdr}
\usepackage[table]{xcolor}

\DeclareMathOperator*{\sumsum}{\sum_\textit{i}\sum_\textit{j}}

\begin{document}
\title{Feature CAM: Interpretability in Image Classification with Improved Visual Attention}
\titlerunning{Feature CAM: Interpretable AI in Image Classification}
%
\author{Frincy Clement\inst{1}\orcidID{0000-0003-0382-653X} \and
Ji Yang\inst{2}\orcidID{0000-0001-9302-4738} \and
Irene Cheng\inst{1}\orcidID{0000-0001-9699-4895 }}
\authorrunning{F. Clement et al.}
%

\institute{Multimedia Research Centre, University of Alberta, 116 St 85 Ave, Edmonton, AB, Canada \\
\email{\{frincy,jyang7,locheng\}@ualberta.ca}\\
\url{http://crome.cs.ualberta.ca/mrc/}}

\maketitle              
\begin{abstract}
Deep Neural Networks have often been called ‘the black box’ because of the complex, deep architecture and non-transparency presented by the inner layers. There is a lack of trust to use Artificial Intelligence in critical and high-precision fields such as security, finance, health, and manufacturing industries. A lot of focused work has been done to provide interpretable models, intending to deliver meaningful insights into the thoughts and behavior of neural networks. 

In our research, we compare the state-of-the-art methods in the Activation-based methods (ABM) for interpreting predictions of CNN models, specifically in the application of Image Classification. We then extend the same for eight CNN-based architectures to compare the differences in visualization and thus interpretability. We introduced a novel technique Feature CAM, which falls in the perturbation-activation combination, to create fine-grained, class-discriminative visualizations. The resulting saliency maps from our experiments proved to be 3-4 times better human interpretable than the state-of-the-art in ABM. At the same time it preserves machine interpretability, which is the average confidence scores in classification.

\end{abstract}

\section{Introduction}
Applications that make use of Artificial Intelligence (AI) have seen an upward trend in the past few years. Due to the advances in this research, various neural network models and architectures were introduced and a number of them have demonstrated accurate and robust performances. While it is a fact that the models can perform complex tasks quite easily, humans are unable to explain how it arrives at a decision. Hence, there is still hesitation to use AI in high-precision applications such as security, health, and manufacturing industries. Because of this, Deep Neural Network (DNN) is often referred to as “the black box”.

\begin{figure}[hbtp]
\centering
{\includegraphics[width=0.9\textwidth]{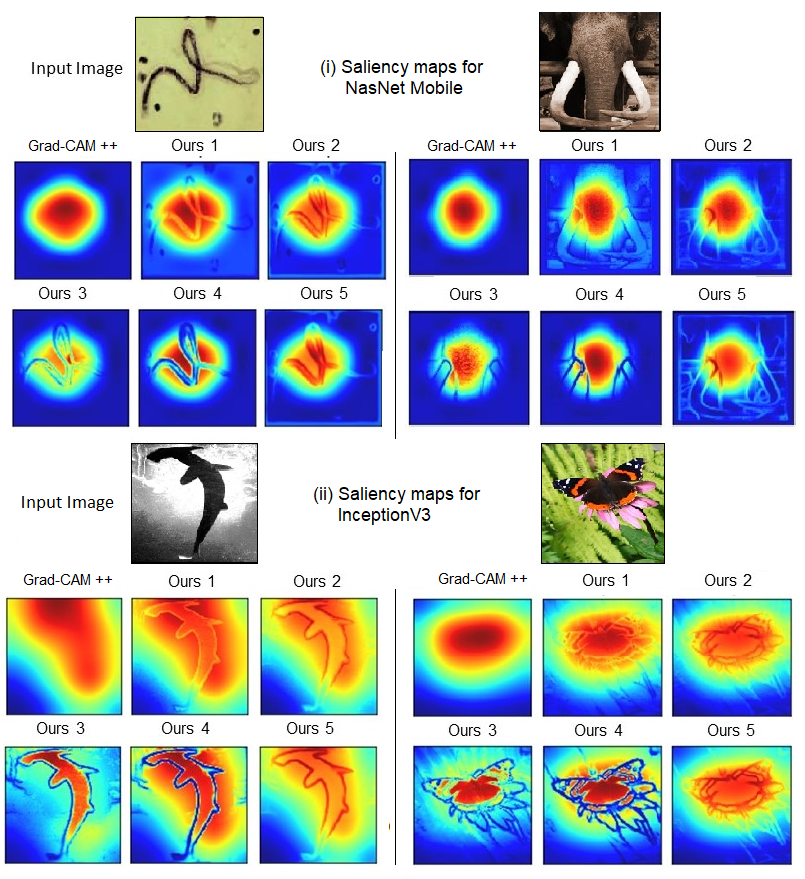}}
\caption{Top 5 Visualizations using Feature CAM in comparison to that of Grad-CAM++ on NasNet Mobile (Top) and InceptionV3 (Bottom). Our methods show visual features of the input image better than Grad-CAM on both classifiers. }
\label{fig:Fig1}
\end{figure}

Interpretable models aim to resolve this issue by providing users with an intuitive method to understand the reason behind the model's prediction. When it comes to the application of image classification, interpretable models create a visualization of features or a map of object localization. It helps to understand where the network is looking in the image, to arrive at that classification. Various approaches have been introduced to tackle this issue which try to interpret the network via image-like representation. But these visualizations are very generic and differ only in localization and does not provide any class-representative features, when visualized for the same classifier.

Our research, which is based on the Activation-Based Methods (ABM) introduces fine-grained details, which are representative of the class the object belongs to, as seen on ``Fig.~\ref{fig:Fig1}''. This makes the visualizations more \textbf{class-discriminative}, which means the resulting visualizations using our approach can better interpret the classification results because of the additional class-representative visual features that were introduced. 

We compared the saliency maps generated by the state-of-the-art techniques in ABM such as Grad-CAM, Grad-CAM ++ and  Smooth Grad-CAM ++ and our proposed technique Feature CAM, on eight different CNN-based classifiers, in addition to VGG16 on which all the previous work has been compared. A qualitative analysis which measured the human interpretability as well as a quantitative analysis which measured the machine interpretability were performed for the evaluation. The results prove that our saliency maps were 3-4 times better interpretable than the existing techniques, while preserving the machine interpretability. 

The remainder of this paper is organized as follows. Section 2 introduces the related works for interpretable models done in the field of image classification. Section 3 details the ABM techniques that were used as baseline for our approach. Our Proposed method is explained in detail in Section 4. Section 5 discusses evaluation methodology and results. Finally, Section 6 gives the conclusion and future work.

\section{Related Work}

The research in the field of interpretability in image classification can be broadly divided into three categories, based on the type of operation they perform to make the model interpretable.

\subsection{Backpropagation Based Methods (BBM)}
This family of methods uses the backpropagation of error signals as a way of measuring the importance of each pixel in an image. It creates an importance map that is overlaid on the original image to understand where the network is looking, as a way of interpreting their decision.

Simonyan et al. \cite{b2}, proposed GRAD, first of its kind, which introduced two techniques for visualization, based on the gradient of class scores with respect to the input image. They generated a synthetic image which maximized the class score and computed a class saliency map to represent pixel-wise importance for each input.

Zeiler et al. \cite{b7} added a Deconvnet layer to each convolutional layer to visualize the features, which were more discriminative and invariant as they go to the higher layers, and called it the Deconvolution. Springenberg et al. \cite{b6}, introduced Guided Backpropagation, a variant of deconvolution approach which helped to visualize features, by adding an additional guidance signal to the backpropagation.

Excitation Backpropagation \cite{b14} introduced the concept of contrastive attention to make the top-down attention maps more discriminative and to identify the task-relevant neurons in the network. Other methods such as Layer wise Relevance Propagation (LRP) \cite{b8}, DeepLift \cite{b11} also employ a top-down propagation for visual representation.

The approach by Sundararajan et al. \cite{b10} makes few calls to the gradient operator, to attribute the predictions of a deep neural network to its input features. Smoothgrad \cite{b9} introduced a new concept of adding noise - to remove noise and to sharpen the gradient-based sensitivity maps by applying a smoothing effect. 

While BBMs are computationally fast, the importance maps produced by them are of low quality and they are not able to provide clear object localization compared to PBM and ABM discussed in the next two sections.

\subsection{Perturbation Based Methods (PBM)}
In this approach, some form of perturbation is performed on the input image to monitor the change in the model’s prediction, in order to understand the regions importance for each prediction. 

Zeiler et al \cite{b7} devised a method, which occluded the input image with a gray square, from left to right and top to bottom, while observing the changes in class probabilities to understand the importance of a feature.

Following Zeiler’s work, there were a few approaches developed for interpretation using PBM, such as discrepancy maps by Zhou et al. \cite{b15}, LIME by Ribeiro et al. \cite{b16}, Regional multi-scale approach by Seo et al. \cite{b17} and RISE by Petsiuk et al. \cite{b18} However, a major drawback of PBM is that they are often time-consuming and fail to generate fine-grained explanations.

\subsection{Activation Based Methods (ABM)}
Activation-based methods create visualization called saliency maps, by using a linear combination of activations from convolutional layers of an image classifier. This approach is comparatively newer and is the focus of our research. The latest ABM are capable of producing saliency maps with exceptional localization. When combined with other methods like Guided-Backpropagation which is a BBM, the ABM provides fine-grained class-discriminative visual explanations on top of localization, an example for this is Guided Grad-CAM. In our research, we have taken a similar approach of combining ABM with perturbed version of original input to create fine-grained saliency maps with good localization capability.

\section{ABM as Baseline For Our Approach}

First of its kind, Class Activation Mapping (CAM) technique was introduced in 2016 by Zhou et al.\cite{b1} Class activation maps indicate the class discriminative image regions used by CNN models to identify a class of the object. Saliency maps are generated by performing global average pooling on the penultimate layer feature maps (convolutional layer just before final fully-connected layer) and by calculating a weighted sum of these values. This sum is used as an input to the final dense layer, which outputs a visual heatmap localizing the object. 

For a given image, \textit{f\textsubscript{k}(x,y)} represents activation of unit \textit{k} in the last convolutional layer at spatial location \textit{(x,y)}, \textit{M\textsubscript{c}(x,y)} is the class activation map for each spatial element and \textit{w\textsuperscript{c}\textsubscript{k}}  is the weight corresponding to class \textit{c} for unit \textit{k}. The class activation map \textit{M\textsubscript{c}} is given by:

\begin{equation}
M_c(x,y) = \sum_k w^c_k f_k(x,y)\label{eq:1}
\end{equation}

Thus the given equation depicts the importance of activation at spatial location \textit{(x,y)}, which finally leads to the object being classified as class \textit{c}. However, the major drawback of the method is that they were not generalized to be applied to all CNN-based classifiers. 

Selvaraju et al. \cite{b3}, introduced Grad-CAM which added GAP (Global Average Pooling) on the gradient of class scores w.r.t feature maps as weights in the linear combination of activations, rather than weights in the network, which resulted in a better discriminative object localization and visualization. As per Eqn.\eqref{eq:2} this new weight \textit{ \textalpha \textsuperscript{c}\textsubscript{k}} is obtained by applying GAP on gradient of class scores \textit{Y\textsuperscript{c}} w.r.t feature maps \textit{A\textsuperscript{k}} of convolutional layer.

\begin{equation}
\alpha^c_k = \frac{1}{Z} \sumsum^{GAP} \frac{\partial Y^c}{\partial A^k_{ij}} \label{eq:2}
\end{equation}

The class discriminative localization heatmap \textit{L\textsuperscript{c}} for Grad-CAM is generated by applying ReLU on the weighted combination activation units \textit{A\textsuperscript{k}}.   

\begin{equation}
L^c_{Grad-CAM} = ReLU \left(\sum_k \alpha^c_k A^k \right) \label{eq:3}
\end{equation}

Even though, Grad-CAM produced better results and good accuracy rate compared to CAM, it failed to classify images with multiple occurrences of the same class.

To tackle the above limitation, Grad-CAM++ \cite{b4} introduced by Chattopadhay et al., used a pixel-wise weighted combination of positive partial derivatives of the feature maps from the penultimate layer. This technique required only single backward pass for computing the linear combination of activations and hence, was computationally efficient compared to the prior methods. \textit{w\textsuperscript{c}\textsubscript{k}} is the new weight used in Grad-CAM ++ which is obtained by taking weighted average of pixel-wise gradients. 

\begin{equation}
w^c_k = \sumsum \alpha^{kc}_{ij}.relu \left( \frac{\partial Y^c}{\partial A^k_{ij}} \right) \label{eq:4}
\end{equation}

\textit{\textalpha\textsuperscript{kc}\textsubscript{ij}}  is weighted coefficients of gradients at pixel \textit{(i,j)} for class \textit{c} at unit \textit{k}.

\begin{equation}
\alpha^{kc}_{i,j} = \frac { \frac{\partial^2Y^c}{(\partial A^k_{i,j})^2}} {2\frac{\partial^2Y^c}{(\partial A^k_{i,j})^2} + \sum_a\sum_b A^k_{a,b} \frac{\partial^3Y^c}{(\partial A^k_{i,j})^3}}  \label{eq:5}
\end{equation}

To improve the quality of visualization by Grad-CAM++, Smooth Grad-CAM ++ was introduced by Omeiza et al. \cite{b5} It addressed the need for getting enhanced visual sharpness in the activation maps. Smooth Grad-CAM++, combined Smooth Grad (which is a BBM) with Grad-CAM++, to get a smoothening effect which resulted in improved visualization and interpretability as per the author. In this method heatmaps were sharpened by taking random instances adjacent to the input \textit{x} and taking the mean of all saliency maps where \textit{n} is the number of sample images, \textit{N(0, \textsigma\textsuperscript{2})} represents Gaussian noise with standard deviation \textsigma. \textit{M\textsubscript{c}} is new saliency map after applying smoothening effect.

\begin{equation}
M_c(x) = \frac {1}{n}\sum_1^n M_c (x+ N(0,\sigma^2)) \label{eq:6}
\end{equation}

The resulting image covers a larger area while preserving the localization. But, our quantitative study on the saliency maps generated by the three techniques on VGG16 indicates that the Grad-CAM++ gives the best confidence scores and is considered to be the state-of-the-art in ABM. Our motivation was to improve the interpretability obtained through the ABM by making it more fine-grained and class-discriminative along with its localization capabilities, thereby increasing confidence of the application users in Artificial Intelligence. In our research, we investigated several approaches by enhancing state-of-the-art ABM methods, leading to a better visualization aspect of the interpretable models for image classification. 

\section{Proposed Method: Feature CAM}
While the various methods provide good interpretability, there is usually a trade-off between localized vs fine-grained explanation. Hence, we propose to devise a method that combines both localization and fine-grained details. The motivation for our research was to improve the human interpretability of the saliency maps generated by the Grad-CAMs while preserving the machine interpretability. 

\textit{\textbf{Human interpretability}} is measured by the improvement in human interpretation provided by the saliency maps, evaluated through a qualitative study. \textbf{\textit{Machine interpretability}} is measured by comparing the average confidence scores of classification provided by the original pixels represented by Feature CAM saliency maps, in comparison to the original image as well as the saliency maps generated by Grad-CAMs.

We started our research by extending the existing ABM techniques to be generalized on multiple CNN-based architectures. These techniques provide either fine-grained or otherwise localized interpretation, but not both. We performed a series of different experiments to create novel technique which can improve the interpretability of saliency maps generated by CNN-based classifiers. Our proposed technique falls in the category of a combination of perturbation and activation-based methods, as one of the input is a feature descriptor, in other words, a perturbed version of the input image.

For our proposed technique Feature CAM, we investigated optimal feature descriptors of the original input image, to be combined with the saliency maps of Grad-CAMs. We selected three edge-based feature descriptors using Holistically Nested Edge Detection (HED) \cite{b21}. It is a deep learning model used for detecting edges of an image. The feature descriptor is then used to create two feature enhanced input images at specific blending ratios 2:1 and 1:1, which become second and third input. The inputs used in all of the experiments are as follows: (i) Feature only, (ii) Feature enhanced original image version 1 and (iii) Feature enhanced original image version 2 as shown below.
 
 \begin{figure}[hbtp]
\centering
{\includegraphics[width=0.6\textwidth]{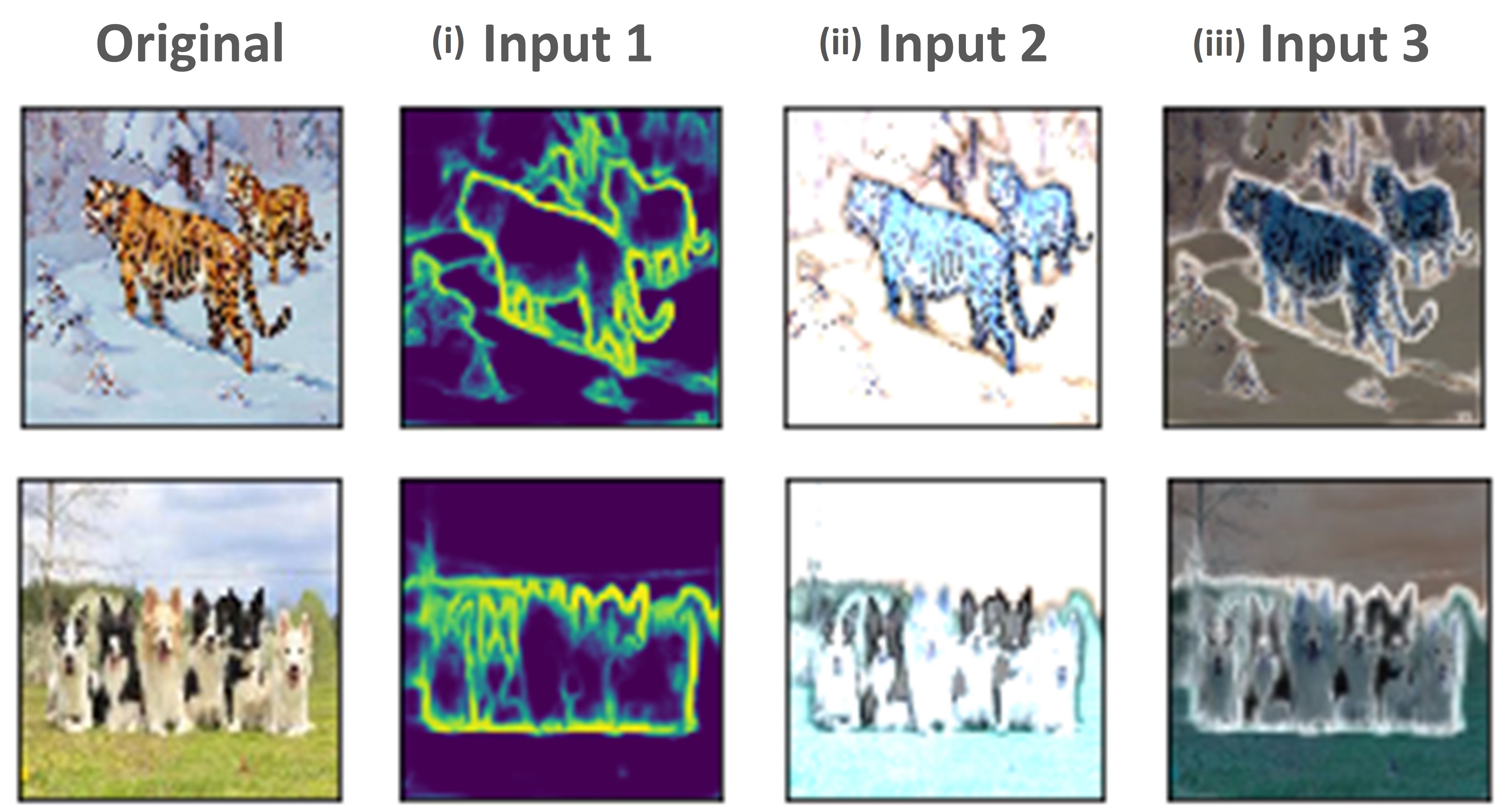}}
\caption{Different versions of perturbed inputs used for performing experiments}
\label{fig2}
\end{figure}

 We combine the perturbed versions of input image with the saliency maps by Grad-CAMs using three experiments, such that it adds fine-grained details of the class in the saliency maps without loosing the localization. An initial evaluation was done by comparing the results using 8 sample inputs on nine classifiers to qualitatively understand human interpretability. The same experiments were performed on selected classifiers (two lighter models VGG16 and NasNetMobile, two heavier models InceptionV3 and ResNet) to quantitatively understand machine interpretability. This resulted in selecting top 5 out of the nine combinations (three inputs and three experiments) across all nine classifiers. 
 
 \begin{figure}[hbtp]
\centering
{\includegraphics[width=0.9\textwidth]{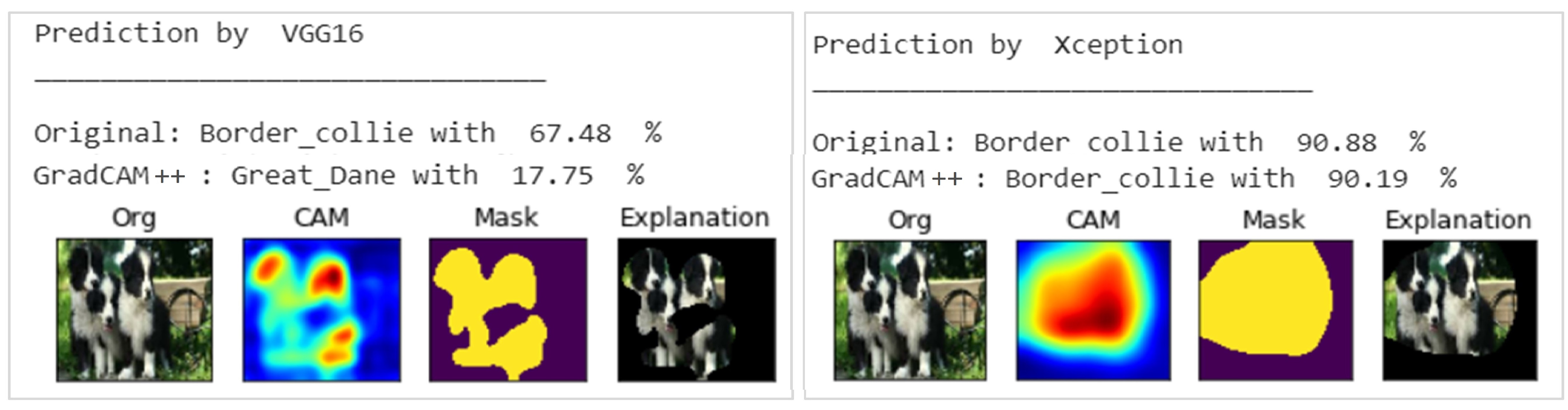}}
\caption {Stages of creation of an explanation map. L to R: Original Image, Saliency Map, Mask and Explanation Map across classifiers VGG16 and Xception on Grad-CAM++}
\label{fig:fig3}
\end{figure}

 For performing this evaluation, we created \textbf{explanation maps} for the saliency maps generated by Feature CAM and Grad-CAMs as seen on ``Fig.~\ref{fig:fig3}''. First, binary thresholding was applied on the saliency maps to extract its mask, where 80 percentile of intensities were passed to the foreground. The pixels of the original image on the corresponding spatial location are then copied on the mask to get the explanation map. These explanation maps by Feature CAM are then passed to the classifiers and compared with the results given by passing original image as well as explanation maps of Grad CAMs.

\begin{figure}[hbp]
\centering
{\includegraphics[width=\textwidth]{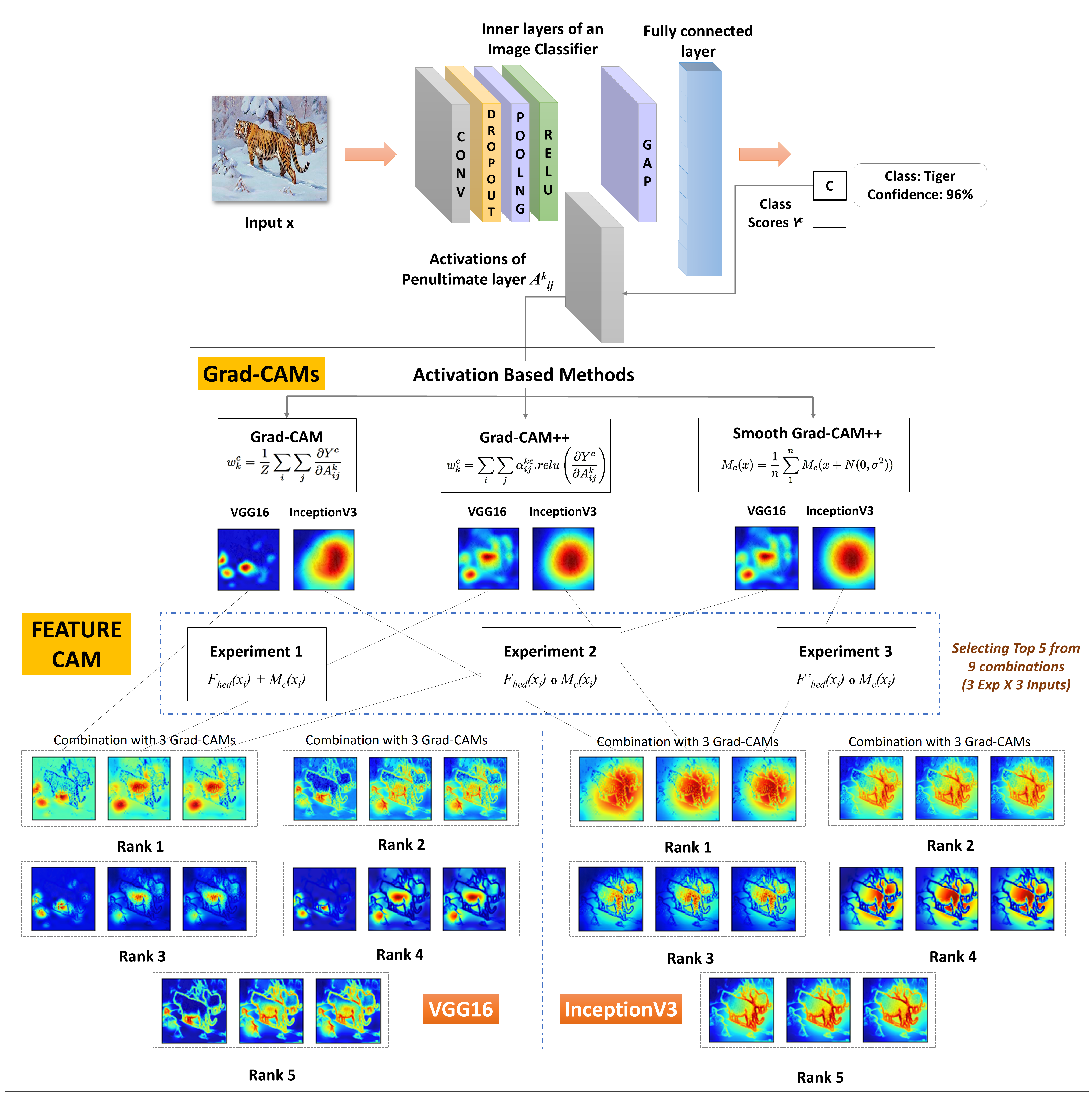}}
\caption{A processing pipeline explaining the intuition behind Feature CAM. The top half gives an overview of the Grad-CAMs starting from extracting penultimate activations before Global Average Pooling and performing linear combinations to create saliency maps. The bottom half shows Feature CAM, its experiments and the top 5 visualizations generated using a lighter classifier (VGG16) and larger classifier (Inception V3) }
\label{fig4}
\end{figure}
 
 We used three metrics for initial evaluation. They are: (i) Percentage of number of right classifications given by Feature CAM compared to that of original image; (ii) Percentage of number of times there is an increase in confidence score by Feature CAM from that of the original image and (iii) Percentage number of times there is an increase in confidence score by Feature CAM, compared to the existing Grad-CAM techniques. Following the initial evaluation, we validate the findings by evaluating them on a larger dataset, for which the results are presented in the next section. A comprehensive visualization of the implementation pipeline of Feature CAM can be seen in Fig.~\ref{fig4}.
 
 \medskip
\noindent Below, we present the experiments and the results of initial evaluation.\\

\noindent\textbf{Experiment 1: Pixel-wise addition of perturbed images with Grad-CAMs} : Pixel-wise addition was performed between the perturbed images and the saliency maps of three Grad-CAMs. The three input versions were weighted by a threshold \textit{k} between 0 and 0.5 such that the resulting visualizations still preserve the heatmap but add recognizable features inside it. For our experiment, we have kept the value of \textit{k} constant at 0.25. 

\begin{equation}
k.F_{hed}(x_i)+ M_c(x_i)  \label{eq:11}
\end{equation}

\textit{F\textsubscript{hed}} denotes the three input versions using HED and \textit{M\textsubscript{c}} denotes the saliency map produced by Grad-CAMs.

This experiment yielded the best results on all three metrics for input 2. It was followed by input 3 and input 1, with varied performance on all metrics. So, all of the 3 inputs combinations using experiment 1 was selected in the Top 5 combinations.

\noindent\textbf{Experiment 2: Pixel-wise multiplication of perturbed images with CAMs} : In this experiment,  the perturbed images were multiplied with the saliency maps of three Grad-CAMs. 

\begin{equation}
F_{hed}(x_i) \circ M_c(x_i)  \label{eq:12}
\end{equation}

This experiment did not yield as expected in any of the three metrics. Hence no result from this experiment was selected.

\noindent\textbf{Experiment 3: Pixel-wise multiplication of complemented perturbed images with CAMs} : In this experiment, the complement of perturbed images were multiplied with the saliency maps of three Grad-CAMs, which resulted in a completely different set of results. 

\begin{equation}
F_{hed}'(x_i) \circ M_c(x_i)  \label{eq:13}
\end{equation}

The experiment yielded better results for input 3 on all the metrics and for input 1 for two metrics. Hence, two of the input combinations were selected from this experiment into the top 5 combinations. 
\begin{figure}[hbp]
\centering
{\includegraphics[width=\textwidth]{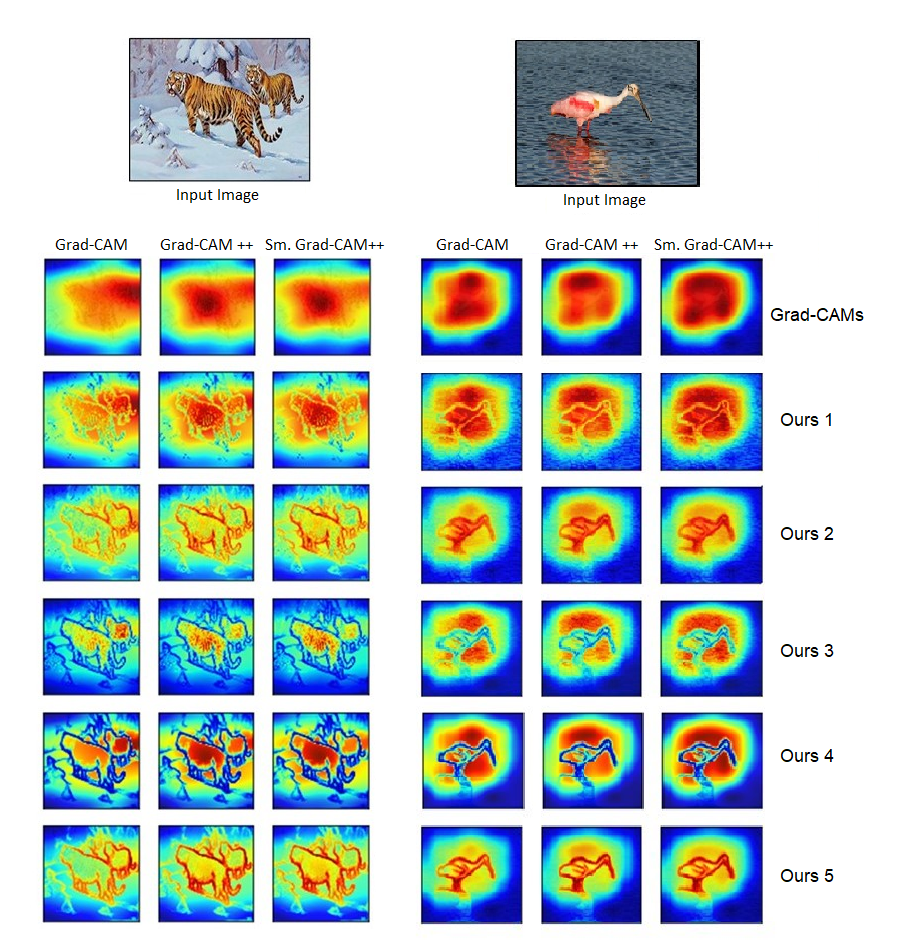}}
\caption{Top five visualizations using Feature CAM on larger classifier - Inception ResNetV2 (left) and lighter classifier - MobileNet(right) in comparison to Existing Grad-CAM techniques }
\label{fig5}
\end{figure}



\section{Empirical Evaluation of Results}

The top 5 saliency maps selected from the initial analysis are then used for performing qualitative analysis on a group of 25 individuals and quantitative analysis is performed on a larger data set. We chose a subset of images from 30 ImageNet classes to be evaluated on two smaller classifiers (VGG16 and NasNet Mobile) and a larger classifier (InceptionV3).

\subsection{Qualitative Analysis }

Since the quality of visualization is subjective, there is a need for defining human interpretability in terms of some metrics which would help us compare the results. We performed qualitative analysis for Feature-CAM by conducting a survey in a controlled environment with 25 participants, who are aged between 25 and 30 years and are graduate students in Computing Science.  The survey aimed at studying two metrics: (i) Human Faith \% and (ii) Interpretability Index.

(i) \textbf{Human Faith \%}: The problem that we were addressing was to improve the human faith in AI by providing better human interpretable visualizations. Hence, understanding the increase in human faith was imperative. To measure human faith \% we provided the users with one saliency map at a time and asked them to evaluate if they can interpret the original image from a small pool of images by just using the saliency map. Using this, we measure human faith as \textit{the percentage of the number of users, who could interpret the class from a saliency map, when a set of original images were given as the reference}. 

As given in Table \ref{tab:table3}, human faith \% in Feature CAM is twice better than Grad-CAMs for top 3 results, followed by the remaining two as well.

\medskip
(ii) \textbf{Interpretability Index}: To compare the interpretability of Feature CAM visualizations with existing Grad-CAMs, we used Interpretability Index. It is measured by \textit{the average rating given by the users on a scale of 0-6 in terms of how best they think the saliency map interprets the original image}. 
As given in the Table \ref{tab:table3}, interpretability index for Feature-CAM is 3-4 times better than existing Grad-CAMs. This means that Feature-CAM visualizations are 3-4 times better human interpretable than existing Grad-CAMs.

\begin{table}[hbtp]
  \begin{center}
    \caption{Results of Qualitative Analysis for Feature CAM}
    \label{tab:table3}
    \begin{tabular}{|l|c|c|c|c|c|c|} 
      \hline
      {} & { Grad-CAMs } & { Ours 1 }& { Ours 2 }& { Ours 3 } & { Ours 4 } & { Ours 5 }\\
      
      \hline 
      Human Faith \% & 48\% & 96\% & 92\% & 92\% & {80\%} & 76\%\\
      \hline
      Interpretability Index & 1.02 & 4.05 & 3.63 & 3.47 & 3.24 & 3.30 \\
      
      \hline
    \end{tabular}
  \end{center}
\end{table}

\subsection{Quantitative Analysis}
To understand the machine interpretability, we need to quantitatively measure the confidence scores of explanation maps generated by Feature CAMs on a larger ImageNet dataset. The explanation maps by Feature CAM and existing techniques were passed through the same classifier that generated them for prediction and the confidence scores were calculated. 

The input dataset we have used for this analysis is a subset of ImageNet with 30 classes. This analysis was performed for one stronger classifier (InceptionV3) and two lighter classifiers (VGG16 and NasNet Mobile). We measure the performance of explanation maps using two metrics: Right Classification \% and Confidence Scores

(i) \textbf{Right Classification (\%)}: This metric compares the number of times the explanation maps by Feature CAM are rightly classified compared to the original classification by Grad-CAMs. As per the results given on Table \ref{tab:table4}, for InceptionV3, three versions of Feature CAM (1,2 \& 5) performs similar to the Grad-CAMs. For a smaller classifier like VGG16, the \% of right classification has improved really well compared to all the Grad-CAM techniques, especially for Grad-CAM. Similarly, for NasNet Mobile, Feature CAM has improved the \% of right classification by a large margin, especially for Grad-CAM ++ and Smooth Grad-CAM++.

\begin{table}[hbp]
  \begin{center}
    \caption{Comparison of \% of Right Classification: Grad-CAMs Vs Feature CAM (Ours) on ImageNet 30 classes}
    \label{tab:table4}
    \begin{tabular}{l|l|c|c|c|c|c|c} 
      \hline
      {Classifier} & {Grad-CAM Versions} & {Grad-CAM} & {Ours 1}& {Ours 2}& {Ours 3} & {Ours 4} & {Ours 5}\\
      
      \hline 
      Inception & Grad-CAM  & 93\% & 90\% & 92\% & {92\%} & 85\% & \textcolor{red} {93\%}\\
      {V3} & Grad-CAM ++  & 95\% & \textcolor{red} {95\%} & \textcolor{red} {95\%} & {88\%} & 80\% & \textcolor{red} {95\%}\\
      {} & SMGrad-CAM ++   & 97\% & 96\% & 94\% & {87\%} & 77\% & 95\%\\
      \hline
     VGG16 & Grad-CAM  & 52\% & \textcolor{red} {78\%} & \textcolor{red} {81\%} & 51\% & 50\% & \textcolor{red} {66\%}\\
      {} & Grad-CAM ++  & 80\% & \textcolor{red} {88\%} & \textcolor{red} {86\%} & \textcolor{red} {84\%} & \textcolor{red} {81\%} & \textcolor{red} {84\%}\\
      {} & SMGrad-CAM ++ & 80\% & \textcolor{red} {88\%} & \textcolor{red} {86\%} & \textcolor{red} {84\%} & \textcolor{red} {81\%} & \textcolor{red} {84\%}\\
      \hline
     NasNet  & Grad-CAM  & 85\% & \textcolor{red} {92\%} & \textcolor{red} {95\%} & \textcolor{red} {88\%} & \textcolor{red} {87\%} & \textcolor{red} {90\%}\\
     Mobile & Grad-CAM ++  & 75\% & \textcolor{red} {88\%} & \textcolor{red} {93\%} & \textcolor{red} {79\%} & \textcolor{red} {79\%} & \textcolor{red} {83\%}\\
      {} & SMGrad-CAM ++ & 73\% & \textcolor{red} {90\%} & \textcolor{red} {93\%} & \textcolor{red} {79\%} & \textcolor{red} {79\%} & \textcolor{red} {89\%}\\
      \hline
    \end{tabular}
  \end{center}
\end{table}

\noindent \textit{Inference: Feature CAM improves the percentage of right classification for lighter/ smaller classifiers by a large margin compared to Grad-CAM techniques. For larger/stronger classifiers, Feature CAM does not improve the values compared to Grad-CAM techniques, but attempts to preserve it even after adding the fine-grained details of the class.}

\medskip

(ii) \textbf{Confidence Score (\%)}: To understand the machine interpretability, the average confidence scores of explanations by visualizations of Grad-CAMs and novel Feature CAM were compared with that of the original image. As per the results given in Table \ref{tab:table5}, for larger classifiers like InceptionV3 as well as lighter classifiers like VGG16 and NasNet Mobile, the Feature CAM preserves the confidence scores and hence the machine interpretability, when compared to Grad-CAMs as well as the original image. 

\begin{table*}[hbtp]
  \begin{center}
    \caption{Comparison of Avg. Confidence Scores: Original Vs Grad-CAMs(G-CAMs) Vs Feature CAM (Ours) on ImageNet 30 classes}
    \label{tab:table5}
    \begin{tabular}{l|l|c|c|c|c|c|c|c} 
      \hline
      {Classifier} & {G-CAM Versions} & {Original} & {G-CAMs} & {Ours 1}& {Ours 2}& {Ours 3} & {Ours 4} & {Ours 5}\\
      
      \hline 
      Inception & Grad-CAM & 97\% & 97\% & \textcolor{red} {98\%} & \textcolor{red} {98\%} & \textcolor{red} {97\%} & \textcolor{red} {97\%} & \textcolor{red} {97\%}\\
      {V3} & Grad-CAM ++ & {} & 96\% & \textcolor{red} {97\%} & \textcolor{red} {97\%} & \textcolor{red}{96\%} & 95\% & \textcolor{red}{96\%}\\
      {} & SMGrad-CAM ++ & {}  & 96\% & \textcolor{red}{96\%} & \textcolor{red} {98\%} & \textcolor{red}{96\%} & \textcolor{red} {97\%} & \textcolor{red}{97\%}\\
      \hline
     VGG16 & Grad-CAM & 86\% & 82\% & \textcolor{red}{83\%} & \textcolor{red} {86\%} & \textcolor{red}{84\%} & \textcolor{red}{84\%} &\textcolor{red} {84\%}\\
      {} & Grad-CAM ++ & {} & 88\% & \textcolor{red} {88\%} & \textcolor{red} {89\%} & \textcolor{red} {88\%} & 85\% & 86\%\\
      \hline
    NasNet & Grad-CAM & 87\% & 78\% &\textcolor{red} {85\%} & \textcolor{red}{84\%} & \textcolor{red}{78\%} &\textcolor{red} {80\%} & \textcolor{red}{82\%}\\
    Mobile & Grad-CAM ++ & {} & 73\% & \textcolor{red} {84\%} & \textcolor{red} {85\%} & \textcolor{red} {78\%} & \textcolor{red} {77\%} & \textcolor{red} {83\%}\\
      {} & SMGrad-CAM ++ & {}  & 75\% & \textcolor{red} {83\%} & \textcolor{red} {84\%} & \textcolor{red}{77\%} & \textcolor{red} {77\%} & \textcolor{red} {81\%}\\
      \hline
    \end{tabular}
  \end{center}
\end{table*}
\noindent\textit{Inference: Feature CAM preserves the confidence scores and thus machine interpretability for both lighter/smaller classifiers as well as for larger classifiers. }\\

In this section, we performed various analysis to compare the saliency maps by Feature CAM with the latest Grad-CAM techniques\cite{b3,b4, b5}. We demonstrated that our technique Feature CAM were able to produce fine-grained as well as class-discriminative visualizations, which are 3-4 times better human interpretable than the state-of-the-art techniques. Moreover, the quantitative analysis proved that Feature CAM preserves and at times outperforms the machine interpretability compared to Grad-CAMs and the original image. 

\section{Conclusion and Future Work}
Our research was focused on increasing the human faith in AI by improving the interpretability of Deep Neural Networks. Our research has the following contribution:

We created a novel technique called Feature CAM which improved the localized saliency maps generated by Grad-CAMs to make it fine-grained and much more class-discriminative. The resulting top 5 versions of saliency maps generated from our technique are 3-4 times better than the state-of-the-art methods based on human interpretation. This was measured through a comprehensive qualitative study with 25 participants. The original pixels represented by the generated saliency maps also preserves the machine interpretability for classification, in comparison to the original image and the saliency maps by ABM techniques such as Grad-CAMs. Machine interpretability was evaluated by comparing the average confidence scores of a subset of images from 30 ImageNet classes for a smaller classifier (VGG16) and a larger classifier (InceptionV3). Our techniques presented similar confidence scores and occasionally outperformed the scores from original images as well.

A limitation of Feature-CAM is that it depends on Grad-CAMs for the localization. In the future, we would like to improve our work by creating a new baseline for localization, which performs well for larger as well as smaller classifiers, so the localization of Feature CAM will not have to depend upon Grad-CAMs.

\textbf{Acknowledgments.} The authors would like to thank Dr. Lihang Ying and the Multimedia Research Centre (MRC), University of Alberta for their support and motivation. We would also like to acknowledge some of initial contributions in literature review given by Kirtan Shah and Dhara Pancholi.


\begin{thebibliography}{00}
\bibitem{b1} Zhou, Bolei, Aditya Khosla, Agata Lapedriza, Aude Oliva, and Antonio Torralba. “Learning Deep Features for Discriminative Localization.” 2016 IEEE Conference on Computer Vision and Pattern Recognition (CVPR) (June 2016). doi:10.1109/cvpr.2016.319.
\bibitem{b2} Simonyan, Karen, Andrea Vedaldi, and Andrew Zisserman. "Deep inside convolutional networks: Visualising image classification models and saliency maps." arXiv preprint arXiv:1312.6034 (2013).
\bibitem{b3} Selvaraju, Ramprasaath R., Michael Cogswell, Abhishek Das, Ramakrishna Vedantam, Devi Parikh, and Dhruv Batra. “Grad-CAM: Visual Explanations from Deep Networks via Gradient-Based Localization.” 2017 IEEE International Conference on Computer Vision (ICCV) (October 2017). doi:10.1109/iccv.2017.74.
\bibitem{b4}Chattopadhay, Aditya, Anirban Sarkar, Prantik Howlader, and Vineeth N Balasubramanian. “Grad-CAM++: Generalized Gradient-Based Visual Explanations for Deep Convolutional Networks.” 2018 IEEE Winter Conference on Applications of Computer Vision (WACV) (March 2018). doi:10.1109/wacv.2018.00097.
\bibitem{b5} Omeiza, Daniel, Skyler Speakman, Celia Cintas, and Komminist Weldermariam. "Smooth Grad-CAM++: An Enhanced Inference Level Visualization Technique for Deep Convolutional Neural Network Models." arXiv preprint arXiv:1908.01224 (2019).
\bibitem{b6} Springenberg, Jost Tobias, Alexey Dosovitskiy, Thomas Brox, and Martin Riedmiller. "Striving for simplicity: The all convolutional net." arXiv preprint arXiv:1412.6806 (2014).
\bibitem{b7}Zeiler, Matthew D., and Rob Fergus. “Visualizing and Understanding Convolutional Networks.” Lecture Notes in Computer Science (2014): 818–833. doi:10.1007/978-3-319-10590-1\_53.
\bibitem{b8}Bach, Sebastian, Alexander Binder, Grégoire Montavon, Frederick Klauschen, Klaus-Robert Müller, and Wojciech Samek. “On Pixel-Wise Explanations for Non-Linear Classifier Decisions by Layer-Wise Relevance Propagation.” Edited by Oscar Deniz Suarez. PLOS ONE 10, no. 7 (July 10, 2015): e0130140. doi:10.1371/journal.pone.0130140.
\bibitem{b9}Smilkov, Daniel, Nikhil Thorat, Been Kim, Fernanda Viégas, and Martin Wattenberg. "Smoothgrad: removing noise by adding noise." arXiv preprint arXiv:1706.03825 (2017).
\bibitem{b10}Sundararajan, Mukund, Ankur Taly, and Qiqi Yan. "Axiomatic attribution for deep networks." Proceedings of the 34th International Conference on Machine Learning-Volume 70. JMLR. org, 2017.
\bibitem{b11}Shrikumar, Avanti, Peyton Greenside, and Anshul Kundaje. "Learning important features through propagating activation differences." Proceedings of the 34th International Conference on Machine Learning-Volume 70. JMLR. org, 2017.
\bibitem{b12}Du, Mengnan, Ninghao Liu, Qingquan Song, and Xia Hu. “Towards Explanation of DNN-Based Prediction with Guided Feature Inversion.” Proceedings of the 24th ACM SIGKDD International Conference on Knowledge Discovery \& Data Mining (July 19, 2018). doi:10.1145/3219819.3220099.
\bibitem{b13}Wagner, Jorg, Jan Mathias Kohler, Tobias Gindele, Leon Hetzel, Jakob Thaddaus Wiedemer, and Sven Behnke. “Interpretable and Fine-Grained Visual Explanations for Convolutional Neural Networks.” 2019 IEEE/CVF Conference on Computer Vision and Pattern Recognition (CVPR) (June 2019). doi:10.1109/cvpr.2019.00931.
\bibitem{b14}Zhang, Jianming, Zhe Lin, Jonathan Brandt, Xiaohui Shen, and Stan Sclaroff. “Top-Down Neural Attention by Excitation Backprop.” Lecture Notes in Computer Science (2016): 543–559. doi:10.1007/978-3-319-46493-0\_33.
\bibitem{b15}Zhou, Bolei, Aditya Khosla, Agata Lapedriza, Aude Oliva, and Antonio Torralba. "Object detectors emerge in deep scene cnns." arXiv preprint arXiv:1412.6856 (2014).
\bibitem{b16}Ribeiro, Marco, Sameer Singh, and Carlos Guestrin. “‘Why Should I Trust You?’: Explaining the Predictions of Any Classifier.” Proceedings of the 2016 Conference of the North American Chapter of the Association for Computational Linguistics: Demonstrations (2016). doi:10.18653/v1/n16-3020.
\bibitem{b17}Seo, Dasom, Kanghan Oh, and Il-Seok Oh. “Regional Multi-Scale Approach for Visually Pleasing Explanations of Deep Neural Networks.” IEEE Access 8 (2020): 8572–8582. doi:10.1109/access.2019.2963055.
\bibitem{b18}Petsiuk, Vitali, Abir Das, and Kate Saenko. "Rise: Randomized input sampling for explanation of black-box models." arXiv preprint arXiv:1806.07421 (2018).
\bibitem{b19}Chollet, Francois. “Xception: Deep Learning with Depthwise Separable Convolutions.” 2017 IEEE Conference on Computer Vision and Pattern Recognition (CVPR) (July 2017). doi:10.1109/cvpr.2017.195.
\bibitem{b20}Szegedy, Christian, Vincent Vanhoucke, Sergey Ioffe, Jon Shlens, and Zbigniew Wojna. “Rethinking the Inception Architecture for Computer Vision.” 2016 IEEE Conference on Computer Vision and Pattern Recognition (CVPR) (June 2016). doi:10.1109/cvpr.2016.308.
\bibitem{b21}Xie, Saining, and Zhuowen Tu. “Holistically-Nested Edge Detection.” 2015 IEEE International Conference on Computer Vision (ICCV) (December 2015). doi:10.1109/iccv.2015.164.
\end{thebibliography}
\end{document}